\documentclass{article}
\usepackage[preprint]{colm2026_conference}

\usepackage{microtype}
\usepackage{hyperref}
\usepackage{url}
\usepackage{booktabs}
\usepackage{amsmath}
\usepackage{listings}
\usepackage[most]{tcolorbox}
\usepackage{tcolorbox}
\usepackage{multirow}
\usepackage{caption}
\usepackage{makecell}
\tcbuselibrary{breakable}
\usepackage{verbatim}
\usepackage{amsthm}
\usepackage{colortbl}
\usepackage{enumitem}
\usepackage{xcolor}

\newcommand{\best}[1]{\textbf{#1}}
\newcommand{\second}[1]{\underline{#1}}

\theoremstyle{definition}
\newtheorem{definition}{Definition}

\usepackage{svg}
\usepackage{wrapfig}
\usepackage[utf8]{inputenc}

\newcommand{\stitle}[1]{\vspace{0.5em}\noindent{\bf #1}}

\usepackage{lineno}

\definecolor{darkblue}{rgb}{0, 0, 0.5}
\hypersetup{colorlinks=true, citecolor=darkblue, linkcolor=darkblue, urlcolor=darkblue}

\title{
Learning Adaptive Reasoning Paths for Efficient\\ Visual Reasoning
}

\author{
    Yixu Huang\textsuperscript{\rm $\spadesuit$} \quad
    Tinghui Zhu\textsuperscript{\rm $\clubsuit$} \quad
    Muhao Chen\textsuperscript{\rm $\clubsuit$} \\
    \textsuperscript{\rm $\spadesuit$}{\small Fudan University} \quad
    \textsuperscript{\rm $\clubsuit$}{\small University of California, Davis} \\
    \; {\small \texttt{yixuhuang23@m.fudan.edu.cn} \quad {\small \texttt{\{thuzhu, muhchen\}@ucdavis.edu}}
    }
}

\begin{document}

\ifcolmsubmission
\linenumbers
\fi

\maketitle

\begin{abstract}

Visual reasoning models (VRMs) have recently shown strong cross-modal reasoning capabilities by integrating visual perception with language reasoning. However, they often suffer from overthinking, producing unnecessarily long reasoning chains for any tasks. We attribute this issue to \textbf{Reasoning Path Redundancy} in visual reasoning: 
many visual questions do not require the full reasoning process. To address this, we propose \textbf{AVR}, an adaptive visual reasoning framework that decomposes visual reasoning into three cognitive functions: visual perception, logical reasoning, and answer application. It further enables models to dynamically choose among three response formats: Full Format, Perception-Only Format, and Direct Answer. AVR is trained with FS-GRPO, an adaptation of Group Relative Policy Optimization that encourages the model to select the most efficient reasoning format while preserving correctness.
Experiments on multiple vision-language benchmarks show that AVR reduces token usage by 50--90\% while maintaining overall accuracy, especially in perception-intensive tasks. These results demonstrate that adaptive visual reasoning can effectively mitigate overthinking in VRMs. Code and data are available at: \url{https://github.com/RunRiotComeOn/AVR}.
\end{abstract}

\section{Introduction}
Visual reasoning models (VRMs; \cite{jian2025look}) have recently demonstrated strong capabilities in multimodal reasoning by integrating visual perception with language reasoning.
These models exhibit emerging reasoning abilities in visually grounded settings, handling compositional~\citep{johnson2017clevr}, multi-hop~\citep{hudson2019gqa}, and logic-intensive tasks~\citep{lu2023mathvista}.
Existing techniques focus on improving the reasoning ability of VRMs.
They borrow the idea from the Reasoning Language Models (RLMs; \cite{jaech2024openai}), adopting Reinforcement Learning with Verifiable Reward (RLVR; \cite{lambert2024tulu,guo2025deepseek}) to enhance reasoning on language \citep{jaech2024openai,guo2025deepseek}.
One line of work decomposes the response of VRMs by separating perception from reasoning~\citep{xia2025visionary,zhang2025r1vl,li2025self} and ask the model to ground its reasoning based on its perceptual outputs.
Through these techniques, VRMs are able to conduct long, complex reasoning with visual inputs.

Despite these performance gains, recent studies show that LRMs frequently suffer from \textbf{overthinking}~\citep{nayab2024concise,chen2025not,sui2025stop,shojaee2025illusion}.
While extended reasoning can improve performance on challenging problems, models often produce unnecessarily long Chain-of-Thought (CoT) reasoning even for simpler ones, leading to excessive token consumption and sometimes even reducing accuracy due to errors introduced in intermediate reasoning steps~\citep{yang2025r}.
Existing attempts to address this issue~\citep{xie2025arm2,wang2025think} largely rely on heuristic strategies derived from RLMs, overlooking the distinct cognitive structure of visual reasoning, particularly the tight relationship between perception and reasoning.

In this work, we argue that a key cause of overthinking in VRMs lies in the cognitive structure of visual reasoning itself. 
Unlike purely textual reasoning, solving a visual question typically involves three cognitive functions: visual perception, logical reasoning, and answer application~\citep{jia2024describe,xia2025visionary}.
However, not all tasks require the full set of these cognitive functions. We observe a phenomenon termed \textbf{Reasoning Path Redundancy}: many questions can be correctly answered without invoking the reasoning function or both perception and reasoning functions. This finding suggests that selectively invoking cognitive functions can reduce redundant reasoning and even improve performance by reducing accumulated errors.

\begin{figure}[t]
\centering
\vspace{-7mm}
\includegraphics[
    width=0.9\linewidth,
    trim=2cm 3cm 4cm 2cm,
    clip
]{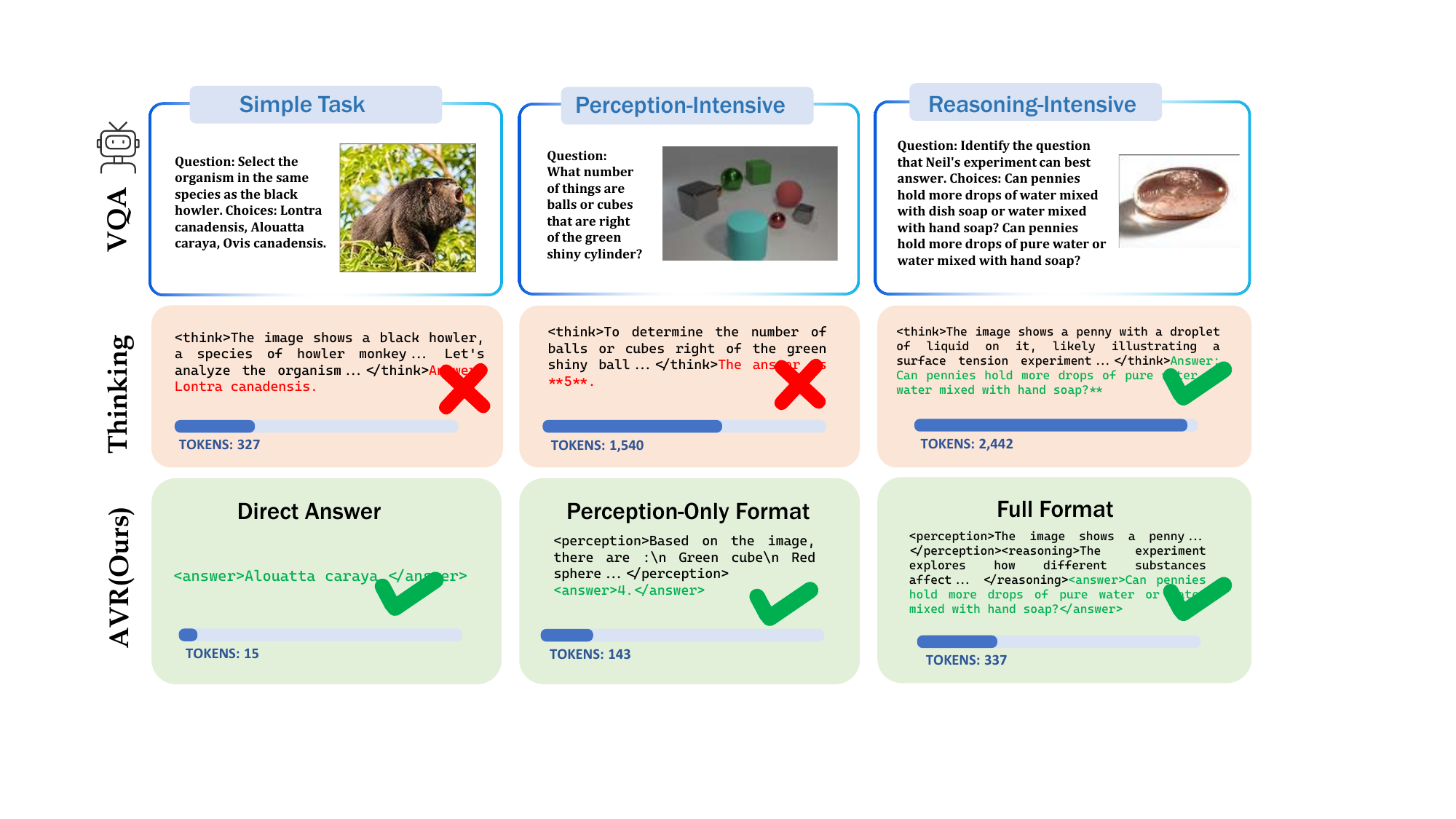}
\caption{
Overview of AVR models.
Top row: Conventional thinking VRMs tend to produce full reasoning paths regardless of task complexity.
Bottom row: AVR decomposes visual reasoning into three cognitive functions and enables adaptive format selection.
}
\vspace{-13pt}
\label{fig:figure1}
\end{figure}

Motivated by this observation, we propose \textbf{AVR}, an adaptive visual reasoning framework that enables VRMs to dynamically allocate reasoning effort based on the task.
AVR decomposes visual reasoning into the previously described cognitive functions and defines three corresponding response formats with different token efficiency: \textit{Full Format}, \textit{Perception-Only Format}, and \textit{Direct Answer}. Rather than enforcing the model to use all cognitive functions, AVR allows the model to adaptively select among these formats according to task requirements.
To enable this adaptive format selection, we adopt a two-stage training pipeline. First, we perform supervised fine-tuning (SFT) on format-annotated data to teach the model to produce responses in different formats. Second, we apply reinforcement learning (RL) using an adapted Group Relative Policy Optimization (GRPO) objective~\citep{shao2024deepseekmath}, which we refer to as Format-Selection GRPO (FS-GRPO), to encourage the model to select the most efficient reasoning format while preserving correctness.

Experiments across multiple benchmarks demonstrate that AVR substantially reduces token consumption while maintaining overall accuracy, outperforming conventional VRMs and prior methods. On perception-intensive tasks like OCRBench, the model predominantly uses direct-answer and perception-only formats, reducing token usage by over 80\% with improved accuracy. On reasoning-intensive benchmarks such as MathVista, it allocates more full-format responses when necessary, reducing token usage while maintaining accuracy.

Our contributions are summarized as follows:

\vspace{-5pt}
\begin{itemize}[leftmargin=1.5em]

\item We identify \textbf{Reasoning Path Redundancy} in VRMs and provide evidence that some visual questions can be solved without reasoning or even through direct answering.

\item We propose \textbf{AVR}, an adaptive visual reasoning framework that combines multi-format SFT with \textbf{FS-GRPO}, an RL objective that jointly optimizes correctness, token efficiency, and format diversity to enable dynamic format selection.

\item AVR achieves \textbf{50--90\% token reduction} across seven benchmarks, while matching or improving accuracy compared to thinking models and other baselines, and generalizes across different model scales and families.
\end{itemize}
\vspace{-5pt}

\section{Related Work}
\vspace{-5pt}
\stitle{Vision-Language Reasoning.}
Recent VRMs have exhibited strong multimodal reasoning capabilities by integrating visual perception with language reasoning. Notably, Multimodal CoT Reasoning~\citep{zhang2023multimodal,zheng2023ddcot,chen2024measuring,zhang2025improve} and Reasoning Decomposition~\citep{jia2024describe,xia2025visionary,zhang2025r1vl,li2025self,huang2025my} have significantly boosted performance on complex tasks. However, they implicitly assume that generating exhaustive reasoning chains is always beneficial~\citep{ma2025boosting,zhang2025llms}. In practice, many visual queries require only minimal perception or direct answers. Therefore, the use of redundant reasoning chains leads to ``overthinking''~\citep{ma2025boosting,zhang2025llms}, which induces computational inefficiency and may even degrade predictive accuracy~\citep{zhu2024unraveling,zhu2025extending,yang2025r,wang2025think}.

\vspace{-5pt}
\stitle{Reinforcement Learning for Reasoning.}
RL has become a primary paradigm for enhancing the reasoning ability of language models~\citep{zhou2025reinforced,zhang2025survey}. Recently, Reinforcement Learning with Verifiable Rewards (RLVR) has emerged as an effective framework for models like OpenAI-o1~\citep{jaech2024openai} and DeepSeek-R1~\citep{guo2025deepseek}, where rewards are computed from verifiable outputs. Specifically, GRPO~\citep{shao2024deepseekmath} offers an efficient RLVR variant. While these methods primarily focus on maximizing reasoning accuracy~\citep{yu2025perception,xia2025visionary,zhang2025r1vl}, they seldom address reasoning efficiency, often resulting in output redundancy~\citep{zhou2025reinforced}.

\vspace{-5pt}
\stitle{Overthinking in Large Reasoning Models.}
Recent studies observe that RLMs frequently encounter severe overthinking when employing CoT reasoning~\citep{nayab2024concise,zhu2024deductive,chen2025not,sui2025stop,shojaee2025illusion}.  
To mitigate this, methods like post-training~\citep{sui2025stop,wu2025arm,tarunokusumo2025boosting} and reasoning trajectory distillation~\citep{sui2025stop,yu2024distilling,dong2025scalable} aim to produce concise yet effective reasoning steps. However, these efforts primarily target text-only models. While preliminary work has begun investigating similar phenomenon in VRMs~\citep{li2025think,xie2025arm2}, methods for mitigating overthinking in VRMs~\citep{xie2025arm2,wang2025think} are largely heuristic, failing to account for the unique structure of multimodal systems. Task-aware approaches that empower VRMs to dynamically adjust their reasoning depth thus remain significantly underexplored.
\vspace{-5pt}
\section{Preliminary}
\label{sec:preliminary}

Based on previous work~\citep{zhang2023multimodal,shao2024visual,aissi2025viper}, the outputs of VRMs follow three structured cognitive functions: 
\vspace{-5pt}
\begin{itemize}[leftmargin=1.5em]
    \item \textbf{Visual Perception}: Visual perception extracts task-relevant information from the input image, including objects, attributes, spatial relations, and textual content.
    \vspace{-2pt}
    \item \textbf{Logical Reasoning}: Logical reasoning performs intermediate inference over perception, including multi-step deduction, comparison, counting, or commonsense reasoning.
    \vspace{-2pt}
    \item \textbf{Answer Application}: Answer application maps the inferred result to the final answer format required by the task.
\end{itemize}
\vspace{-5pt}

However, not all questions require all functions. We refer to this overthinking phenomenon \textbf{Reasoning Path Redundancy}, as defined in Definition~\ref{def:reasoning_path_redundancy}. 

\vspace{5pt}
\begin{definition}[Reasoning Path Redundancy]
\label{def:redundancy}
A reasoning path is \emph{redundant} if a shorter path that contains a subset of its functional components can produce the same correct answer.
\label{def:reasoning_path_redundancy}
\end{definition}

By analyzing 1,000 randomly sampled instances, we observe that VRMs often fail to adaptively select cognitive functions based on task requirements, instead overusing the full set of cognitive functions. In Qwen3-VL-4B-Thinking, 36.7\% of questions can be correctly answered without invoking the reasoning function, and 32.1\% can be solved through direct answering without explicit perception or reasoning (see Appendix~\ref{app:overthinking} for details). 

To further quantify this phenomenon, we introduce an \textbf{Overthinking Score} inspired by the compression ratio from previous works~\citep{song2025reasoning,tang2026towards} as a diagnostic tool to demonstrate the extent of redundancy in reasoning paths. Formally, we define it as the ratio between the token length of the original model response with the correct answer and that of its compressed counterpart:

\vspace{-13pt}
\begin{equation}
\text{Overthinking Score} = \frac{T_{\text{original}}}{T_{\text{compressed}}}
\end{equation}

\begin{wrapfigure}{r}{0.34\textwidth}
\vspace{-20pt}
\centering
\includegraphics[width=\linewidth]{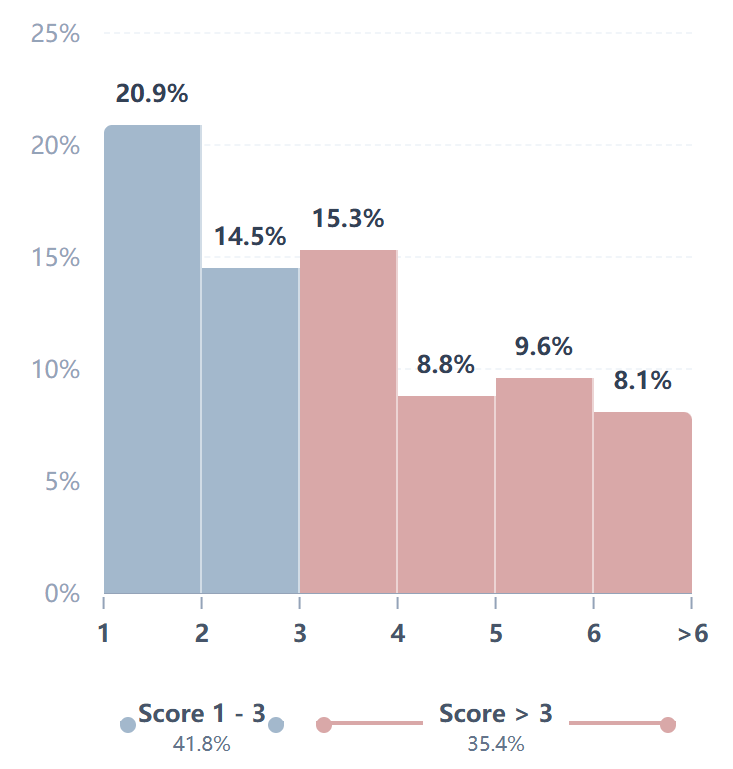}
\caption{Overthinking Score Distribution.
}
\label{fig:overthinking_score}
\vspace{-20pt}
\end{wrapfigure}

where $T_{\text{original}}$ denotes the number of tokens in the original model response, and $T_{\text{compressed}}$ represents the number of tokens in the minimally sufficient response that preserves correctness generated by GPT-4o-mini~\citep{openai2024gpt4omini} (see Appendix~\ref{app:overthinking} for implementation details). Based on this metric, we observe that 35.4\% of instances exhibit an overthinking score greater than 3, while 41.8\% fall within the range of 1 to 3 as demonstrated in Figure~\ref{fig:overthinking_score}. These results further highlight the prevalence and severity of reasoning path redundancy in VRMs.

These results suggest that selectively invoking cognitive functions can preserve response correctness while reducing token consumption. Therefore, adaptive selection of reasoning formats based on task requirements becomes optimal in balancing cost and performance.

\section{Method}
\label{sec:method}

We propose AVR, an RL training framework that enables VRMs to adaptively select efficient formats during visual reasoning process.
Our key idea is to explicitly decompose the reasoning pipeline into three cognitive functions and construct corresponding formats with different levels of reasoning effort. Based on these formats, we train the model to dynamically select the most appropriate reasoning path for each task.

The framework consists of two stages. 
\textbf{Stage 1: SFT for format understanding}. We first fine-tune the model on 11k diverse questions with three-format annotations. This stage enables the model to understand different reasoning formats and produce responses in multiple formats. 
\textbf{Stage 2: RL for efficient format switching}. We then employ FS-GRPO algorithm to train the model to dynamically switch between formats, allowing it to select the most efficient reasoning format based on task requirements while maintaining accuracy.

\subsection{Decomposition of Reasoning Paths}
\label{subsec:decoupling}
We decompose the VRM reasoning process into three cognitive functions as previously stated in Section~\ref{sec:preliminary}: \textbf{Perception}, \textbf{Reasoning}, and \textbf{Answer}. Unlike the monolithic structure of traditional CoT, this decoupled architecture encourages explicit visual grounding and structured logical deduction. Such a design has been shown to improve accuracy on complex visual tasks~\citep{jia2024describe,wang2025your} and provides an explicit mechanism for mitigating overthinking in VRMs (see Appendix~\ref{app:overthinking} for details).

We implement this structure using functional special tokens (\texttt{<perception>}, \texttt{<reasoning>}, and \texttt{<answer>}), which allow us to define three distinct formats ($k$ indexes the format; see Appendix~\ref{app:format_construction} for details):

\vspace{-5pt}
\begin{itemize}[leftmargin=1.5em]
    \item \textbf{Full Format ($k=1$)}: A comprehensive sequence encompassing all components (\texttt{<perception>...<reasoning>...<answer>}).
    \vspace{-2pt}
    \item \textbf{Perception-Only Format ($k=2$)}: A streamlined path combining perception with answer (\texttt{<perception>...<answer>}), bypassing exhaustive reasoning.
    \vspace{-2pt}
    \item \textbf{Direct Answer ($k=3$)}: A minimalist output (\texttt{<answer>}) that targets immediate response efficiency.
\end{itemize}

\vspace{-5pt}
\subsection{Stage 1: SFT for Format Understanding}
In this stage, we leverage SFT as a warm start to introduce the model to the formats it can utilize for solving VQA, using processed data in the formats described in Section~\ref{subsec:decoupling}. After SFT, the model is able to generate all three formats, providing a strategic initialization for subsequent RL training.

\subsection{Stage 2: RL for Efficient Format Switching}
\label{subsec:rl_reward}

While SFT enables the model to follow defined formats, it often fails to dynamically switch between them based on task complexity. To address this, we propose FS-GRPO algorithm that encourages the model to select the most efficient format for a given problem.

For a given prompt $q$, the model generates a group of $G$ outputs $\{o_1, o_2, \dots, o_G\}$. The efficiency and correctness of each output are evaluated through a multi-faceted reward function $r_i$.The total reward $r_i$ for each response $o_i$ is defined as:
\begin{equation}
    r_i =  s_{L, i} \cdot (r_{\text{format}, i} + r_{\text{div}, i}),
    \label{eq:reward_decomposition}
\end{equation}
where $r_{\text{format}, i}$ incentivizes correct answers with efficient formats, $s_{L, i}$ penalizes verbosity, and $r_{\text{div}, i}$ maintains format diversity.

\vspace{-5pt}
\paragraph{Format Reward ($r_{format}$)} 
The base reward evaluates the correctness of the answer while incorporating format-specific rewards:
\begin{equation}
r_{\text{format}} =
\begin{cases}
1 + f_k, & \text{if the answer is correct}, \\
- f_k, & \text{otherwise}.
\end{cases}
\end{equation}
$f_k$ is the specific bonus for format $k \in \{1, 2, 3\}$ as defined in Section~\ref{subsec:decoupling}. 
We use simple fixed values satisfying $f_3 > f_2 > f_1 \geq 0$ to impose a monotonic preference over formats, as higher bonuses encourage shorter responses. The exact magnitudes are not critical (see Section~\ref{subsec:sensitivity} for details); they simply establish a coarse hierarchy that favors more token-efficient formats (direct answers first, followed by perception-only reasoning) while penalizing incorrect answers in compressed formats. This encourages the model to use more concise format when confidence level is high, while falling back to longer formats otherwise.

\vspace{-5pt}
\paragraph{Diversity Reward ($r_{div}$)} 
To prevent \textbf{Format Collapse}~\citep{wu2025arm}, where the model converges exclusively to one particular format (see Section~\ref{subsec:ablation_format_collapse} for details), we introduce a diversity reward that rewards less frequent formats within a sample group. For a group of $G$ responses, let $c_k$ be the count of responses using format $k$. The bonus is defined as:
\begin{equation}
    r_{div} = d(t) \cdot (1 - \text{freq}_k),
\end{equation}
where $\text{freq}_k = \frac{c_k}{G}$. We employ a time-dependent decay factor $d(t) = \frac{1}{2} \left( 1 + \cos\left(\frac{\pi t}{T}\right) \right)$, where $t$ is the current training step and $T$ is the total steps. This decay gradually reduces the influence of the diversity reward, allowing the model to focus on optimal format selection rather than exploration in the later stages of training.

\vspace{-5pt}
\paragraph{Length Scaling Factor ($s_L$)} 
To discourage verbosity, we scale the reward based on the response length $n_{tok}$ relative to a length tolerance $L$:
\begin{equation}
    s_L = \begin{cases} 
    1.0 & \text{if } n_{tok} \le L \\ 
    \frac{L}{n_{tok}} & \text{otherwise}. 
    \end{cases}
\end{equation}
\vspace{-10pt}

Then, the group advantage $\hat{A}_{i,k}$ for all tokens in each output is computed based on the relative rewards within the sampled group to reduce variance:
\begin{equation}
    \hat{A}_{i,k} = \frac{r_i - \text{mean}(\{r_1, r_2, \dots, r_G\})}{\text{std}(\{r_1, r_2, \dots, r_G\})}.
\end{equation}
\vspace{-5pt}

Finally, we optimize the model by maximizing the following FS-GRPO objective, which incorporates the PPO-style clipping~\citep{schulman2017proximal} mechanism and a KL divergence penalty~\citep{ouyang2022training} to ensure training stability:
\vspace{-2mm}

\begin{equation}
\begin{aligned}
\mathcal{J}_{\text{FS-GRPO}}(\theta) &= \mathbb{E} \left[ q \sim P(Q), \{o_i\}_{i=1}^G \sim \pi_{\theta_{\text{old}}}(O|q) \right] \Bigg[ \frac{1}{\sum_{i=1}^G |o_i|} \sum_{i=1}^G \sum_{k=1}^{|o_i|} \bigg\{ \\
& \min \left( \frac{\pi_{\theta}(o_{i,k}|q, o_{i,<k})}{\pi_{\theta_{\text{old}}}(o_{i,k}|q, o_{i,<k})} \hat{A}_{i,k}, \text{clip} \left( \frac{\pi_{\theta}(o_{i,k}|q, o_{i,<k})}{\pi_{\theta_{\text{old}}}(o_{i,k}|q, o_{i,<k})}, 1 - \epsilon, 1 + \epsilon \right) \hat{A}_{i,k} \right) \\
& \quad - \beta \, \text{KL} [\pi_{\theta} \| \pi_{\text{ref}}] \bigg\} \Bigg]
\end{aligned}
\end{equation}
\vspace{-2mm}

\vspace{-10pt}
\section{Experiments}

In this section, we evaluate AVR on a wide range of benchmarks.
Furthermore, we analyze its performance, efficiency, and format selection behavior, compared with other methods.

\subsection{Setup}

\paragraph{Model}
To evaluate the generalizability of AVR framework across varying model scales, we adopt Qwen3-VL-2B/4B/8B-Instruct~\citep{bai2025qwen3} as base models. Experiments on other model families are provided in Appendix~\ref{app:other_model_family}.

\vspace{-5pt}
\paragraph{Training Dataset}
We curate a comprehensive RL training dataset of 44k samples by aggregating 10 diverse datasets: OK-VQA~\citep{marino2019ok}, CLEVR~\citep{johnson2017clevr}, VCR~\citep{zellers2019recognition}, GQA~\citep{hudson2019gqa}, ChartQA~\citep{masry2022chartqa}, OCR-VQA~\citep{mishra2019ocr}, MathVerse~\citep{zhang2024mathverse}, Geometry3K~\citep{lu2021inter}, ScienceQA~\citep{lu2022learn}, and TQA~\citep{kembhavi2017you} 
(see Appendix~\ref{app:dataset_composition}). For the SFT phase, we subsample 11k high-quality cases and generate both full-context and perception-only responses using Gemini-2.5-flash~\citep{comanici2025gemini}. A detailed description of our data processing pipeline is provided in Appendix~\ref{app:sft_data}.

\vspace{-5pt}
\paragraph{Baselines}
Beyond base models, we compare AVR with the official Qwen3-VL-2B/4B/8B-Thinking variants~\citep{bai2025qwen3}, which employ long CoT reasoning. We further include Think-or-Not (TON)~\citep{wang2025think} and ARM2~\citep{xie2025arm2} as representative baselines. TON adopts two response strategies (direct answering and CoT reasoning), allowing the model to bypass explicit reasoning for simple queries while invoking deeper reasoning when necessary. ARM2 provides multiple response formats, including long CoT and short CoT, and employs a RL framework with length-aware optimization to balance reasoning performance and efficiency across different reasoning styles.

All training details are provided in Appendix~\ref{app:implementation_details}.

\subsection{Evaluation}

\paragraph{Benchmarks}
Our selected benchmarks cover three main domains based on task types: perception-intensive, reasoning-intensive, and general. For \textbf{perception-intensive} tasks, we utilize OCRBench~\citep{liu2024ocrbench} and TextVQA~\citep{singh2019towards}, which primarily require fast perception with minimal reasoning. For \textbf{reasoning-intensive} tasks, we include MathVista~\citep{lu2023mathvista} and MathVision~\citep{wang2024measuring} to assess models' ability to solve complex mathematical problems that require intensive reasoning. Finally, for \textbf{general} tasks, we select MMMU~\citep{yue2024mmmu}, MMMU-Pro~\citep{yue2025mmmu}, and MMBench~\citep{liu2024mmbench}, which contain a diverse set of tasks spanning a wide range of difficulties.

\vspace{-5pt}
\paragraph{Metrics}
We evaluate AVR using two metrics: \textbf{accuracy} and \textbf{token consumption} of model responses. Each benchmark reports Accuracy (Acc, \%) and average token usage (Tok.). $\Delta$: Accuracy delta shows absolute improvement ($\text{Acc}_{\text{AVR}} - \text{Acc}_{\text{Thinking}}$); Token delta shows relative reduction percentage ($ (\text{Tok}_{\text{Thinking}} - \text{Tok}_{\text{AVR}}) / \text{Tok}_{\text{Thinking}} \times 100\%$). 

We utilize the toolkit VLMEvalKit~\citep{duan2024vlmevalkit} for evaluation. For non–multiple-choice questions, we employ GPT-4o-mini to evaluate model responses, serving as a proxy for human judgment, with additional manual spot-checking to verify evaluation quality.

\begin{table}[t]
\centering
\small
\vspace{-4mm}
\resizebox{0.97\textwidth}{!}{
\begin{tabular}{llcccccccccccccc}
\toprule
\multirow{2}{*}{\normalsize Size} & \multirow{2}{*}{\normalsize Variant}
& \multicolumn{4}{c}{\normalsize\bf Reasoning-intensive}
& \multicolumn{4}{c}{\normalsize\bf Perception-intensive}
& \multicolumn{6}{c}{\normalsize\bf General} \\

\cmidrule(lr){3-6}
\cmidrule(lr){7-10}
\cmidrule(lr){11-16}

& 
& \multicolumn{2}{c}{MathVista}
& \multicolumn{2}{c}{MathVision}
& \multicolumn{2}{c}{OCRBench}
& \multicolumn{2}{c}{TextVQA}
& \multicolumn{2}{c}{MMMU}
& \multicolumn{2}{c}{MMMU-Pro}
& \multicolumn{2}{c}{MMBench} \\

\cmidrule(lr){3-4}
\cmidrule(lr){5-6}
\cmidrule(lr){7-8}
\cmidrule(lr){9-10}
\cmidrule(lr){11-12}
\cmidrule(lr){13-14}
\cmidrule(lr){15-16}

& 
& Acc & Tok.
& Acc & Tok.
& Acc & Tok.
& Acc & Tok.
& Acc & Tok.
& Acc & Tok.
& Acc & Tok. \\

\midrule

\multirow{5}{*}{2B}
& Thinking      & \second{32.8} & 1570.4 & \best{30.1} & 1622.8 & 78.3 & 478.6 & 73.2 & 329.7 & \best{41.2} & 1289.4 & \best{27.5} & 1346.2 & \second{69.3} & 1421.8 \\
& Instruct      & 27.2 & 900.1 & 21.6 & 932.7 & \second{82.8} & \best{41.1} & 77.1 & \best{45.6} & 27.2 & 807.8 & 22.3 & 692.1 & 67.5 & \best{70.5} \\
& + SFT         & 28.9 & 668.8 & 20.3 & 703.5 & 76.9 & 192.0 & \second{77.5} & 132.5 & 40.4 & 634.5 & \second{27.3} & 604.8 & 66.9 & 212.6 \\
\rowcolor{gray!12}
& \textbf{AVR}           & \best{33.0} & \second{544.7} & \second{29.8} & \second{647.2} & \best{81.4} & 72.1 & \best{79.8} & 46.3 & \best{41.3} & \best{182.9} & 26.9 & \best{322.1} & \best{71.4} & 350.3 \\
& $\Delta$ & +0.2 & 65.3 & -0.3 & 60.1 & +3.1 & 84.9 & +6.6 & 86.0 & +0.1 & 85.8 & -0.6 & 76.1 & +2.1 & 75.4 \\

\midrule

\multirow{5}{*}{4B}
& Thinking      & \best{45.8} & 1724.4 & \best{40.3} & 1823.7 & 81.3 & 476.9 & 77.5 & 501.2 & \second{58.3} & 1332.1 & 36.7 & 1527.9 & \second{75.5} & 1467.4 \\
& Instruct      & 37.9 & 877.8 & 33.3 & 835.7 & \second{84.8} & \best{44.6} & 76.3 & \best{51.9} & 49.1 & 832.7 & 28.4 & 540.2 & 70.9 & \second{136.2} \\
& + SFT         & 38.9 & 693.0 & 33.2 & 793.4 & 82.1 & 157.8 & 75.6 & 204.8 & 47.3 & \second{587.3} & 30.0 & 557.4 & 73.5 & 389.2 \\
\rowcolor{gray!12}
& \textbf{AVR}  & \second{43.1} & \best{488.7} & \second{39.2} & \best{510.9} & \best{85.6} & \second{59.4} & \second{80.3} & \second{63.8} & \best{59.8} & \best{128.7} & \second{38.2} & \second{479.3} & \best{76.4} & \best{224.1} \\
& $\Delta$ & -2.7 & 71.7 & -1.1 & 72.0 & +4.3 & 87.5 & +2.8 & 87.3 & +1.5 & 90.3 & +1.5 & 68.6 & +0.9 & 84.7 \\

\midrule

\multirow{4}{*}{8B}
& Thinking      & \best{51.3} & 1648.5 & \second{48.2} & 1814.7 & 81.9 & 512.1 & \second{79.8} & 486.9 & \second{63.8} & 1421.4 & \second{42.1} & 1455.6 & \best{80.7} & 1384.4 \\
& Instruct      & 45.6 & 933.8 & 43.6 & 897.8 & \second{83.5} & \second{79.4} & 79.8 & \second{40.3} & 58.2 & 554.6 & 37.6 & 725.4 & 75.4 & \second{275.3} \\
& + SFT         & 46.3 & \second{875.2} & 46.0 & \second{854.3} & 83.3 & 202.7 & \best{80.3} & 183.9 & 59.2 & \second{473.7} & 40.4 & \second{701.8} & 77.1 & 283.5 \\
\rowcolor{gray!12}
& \textbf{AVR}  & \second{50.2} & \best{531.6} & \best{49.7} & \best{640.2} & \best{84.2} & \best{58.5} & 79.7 & \best{33.4} & \best{64.0} & \best{335.9} & \best{43.5} & \best{226.3} & \second{79.3} & \best{203.1} \\
& $\Delta$ & -1.1 & 67.7 & +1.5 & 64.7 & +2.3 & 88.6 & +0.4 & 93.1 & +0.2 & 76.4 & +1.4 & 84.4 & -1.4 & 85.3 \\
\bottomrule

\end{tabular}
}
\caption{
Performance of model variants across reasoning-intensive, perception-intensive, and general benchmarks. \textbf{Bold} indicates the best and \second{underline} indicates second best.
}
\vspace{-10pt}
\label{tab:acc_tok_benchs}
\end{table}

\subsection{Results}
Table~\ref{tab:acc_tok_benchs} demonstrates a comprehensive result of our models and some baselines across diverse vision-language benchmarks.

\vspace{-5pt}
\paragraph{Thinking v.s. Instruct.}
Comparing the results of thinking and instruct models in Table~\ref{tab:acc_tok_benchs}, we observe that thinking models degrade by approximately 5\% on perception-intensive benchmarks such as OCRBench and TextVQA, while consuming nearly 10$\times$ more tokens. In these cases, long CoT reasoning tends to induce serious overthinking and may even harm performance. In contrast, for general and reasoning-intensive tasks, thinking models consume moderately more tokens (2–4$\times$) than instruct models while achieving performance gains of around 5\%. This suggests that extended reasoning is beneficial for more challenging problems that require deeper visual reasoning, such as MathVista and MathVision, but also reveals the phenomenon of overthinking.

\vspace{-5pt}
\paragraph{SFT Teaches Formats, FS-GRPO Teaches Selection.}
The SFT model fluctuates in accuracy and increases token usage because it does not learn to select formats. As shown in Table~\ref{tab:acc_tok_benchs}, although SFT learns to generate responses in the three predefined formats, the distribution of formats remains nearly identical across different task types. 
In contrast, after applying FS-GRPO, the model learns to selectively use formats for different task categories (Table~\ref{tab:acc_tok_benchs}). 

\begin{figure}[t!]
\centering
\vspace{-6mm}
\includegraphics[
    width=0.90\linewidth,
    trim=3.5cm 7.2cm 4cm 7.5cm,
    clip
]{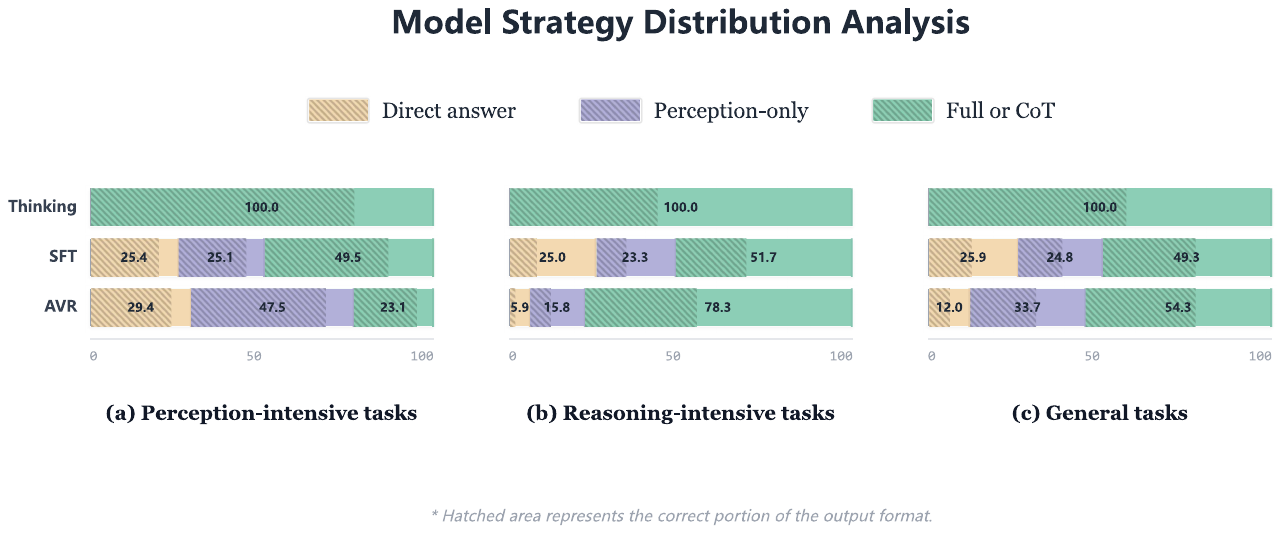}
\vspace{-6mm}
\caption{
Format distribution across different types of tasks using Qwen3-VL-4B.
The hatched areas indicate the proportion of correct answers generated under each format.
}
\label{fig:tasks_format}
\end{figure}

\vspace{-5pt}
\paragraph{AVR v.s. Thinking.}
As shown in Table~\ref{tab:acc_tok_benchs}, compared with conventional thinking models, AVR reduces token consumption while achieving overall performance improvements across all model sizes. Specifically, our model achieves over 80\% token reduction and a 2–4\% accuracy gain on perception-intensive tasks. For reasoning-intensive tasks, the model predominantly adopts the full format as its primary response strategy ($\sim$70\%), enabling sufficient reasoning to derive correct answers while avoiding performance degradation caused by insufficient thinking. Figure~\ref{fig:tasks_format} further confirms that our training framework enables the model to dynamically switch between formats according to task types, effectively achieving adaptive token budgeting. AVR prefers lightweight formats for simple perception-heavy tasks while invoking full reasoning when task complexity requires it.


\begin{wraptable}[11]{r}{0.52\textwidth}
\vspace{-10pt}
\centering
\small
\resizebox{0.48\textwidth}{!}{
\begin{tabular}{lcccccc}
\toprule
\multirow{2}{*}{\normalsize Model} & 
\multicolumn{2}{c}{\normalsize MathVista} & 
\multicolumn{2}{c}{\normalsize OCRBench} & 
\multicolumn{2}{c}{\normalsize MMMU} \\
\cmidrule(lr){2-3} \cmidrule(lr){4-5} \cmidrule(lr){6-7}
& Acc & Tok. & Acc & Tok. & Acc & Tok. \\
\midrule
TON-2B & \second{30.3} & \best{306.1} & 79.8 & \best{58.3} & \second{40.9} & 562.7 \\
ARM2-3B & 29.4 & 636.7 & \second{79.9} & \second{69.2} & 35.2 & \second{331.5} \\
\rowcolor{gray!12}
\textbf{AVR-2B} & \best{33.0} & \second{544.7} & \best{81.4} & 72.1 & \best{41.3} & \best{182.9} \\
\midrule[\heavyrulewidth]
TON-4B & \second{40.3} & 624.2 & \second{79.6} & \second{63.7} & \second{53.6} & 766.7 \\
ARM2-7B & 37.2 & \second{584.3} & 78.8 & 86.9 & 50.4 & 
\second{285.0} \\
\rowcolor{gray!12}
\textbf{AVR-4B} & \best{43.1} & \best{488.7} & \best{85.6} & \best{59.4} & \best{59.8} & \best{128.7} \\
\bottomrule
\end{tabular}
}
\vspace{-5pt}
\caption{Results across reasoning-intensive, perception-intensive, and general benchmarks.}
\vspace{-23pt}
\label{tab:method_comparison}
\end{wraptable}

\vspace{-5pt}
\paragraph{AVR v.s. Other Methods.}
As shown in Table~\ref{tab:method_comparison}, AVR consistently outperforms existing selective reasoning approaches across multiple benchmarks while maintaining competitive token efficiency.  Compared with TON, which adopts a binary strategy between direct answering and full CoT reasoning, AVR achieves higher accuracy on most benchmarks (33.0\% on MathVista and 41.3\% on MMMU). This improvement suggests that the binary reasoning decision in TON may lead to insufficient reasoning on complex tasks.
Compared with ARM2, AVR achieves both higher accuracy and lower token consumption on most benchmarks. 
For instance, on MMMU, AVR-4B reduces the average token usage from 285.0 to 128.7 while improving accuracy from 50.4\% to 59.8\%. 
These results demonstrate that decomposing reasoning paths enables more efficient and adaptive visual reasoning.

\vspace{-5pt}
\paragraph{Dynamic Format Selection Across Tasks.}
\label{para:format_selection}
To understand how AVR allocates reasoning effort, we analyze the distribution of reasoning formats across tasks  (Figure~\ref{fig:tasks_format}). 
For perception-intensive tasks, the model tends to prefer lightweight formats such as direct answers (25.4\%) and the perception-only format (25.1\%), while the use of the full format is reduced after training. The 1--3\% performance improvement is due to shorter formats, which help prevent the model from engaging in unnecessary or error-prone reasoning.
In contrast, for reasoning-intensive tasks, AVR models shift strongly toward the full format (78.3\%), indicating that deeper reasoning is invoked when necessary. For general tasks, the model adopts a mixed strategy with moderate use of the full format while still achieving improvements in accuracy. Detailed analysis of format selection and case studies is provided in Appendix~\ref{app:fotmat_selection} and \ref{app:case_study}. 
These results suggest that AVR adaptively allocates reasoning effort based on task requirements rather than uniformly applying long reasoning chains.

\section{Analysis}

We conduct detailed analyses to better understand the design and behavior of AVR. Specifically, we examine the effects of format composition, diversity reward, and key hyperparameters on both performance and efficiency.

\begin{figure}[t]
\centering
\vspace{-5mm}
\includegraphics[
    width=0.90\linewidth,
    trim=1.5cm 13cm 1.5cm 1cm,
    clip
]{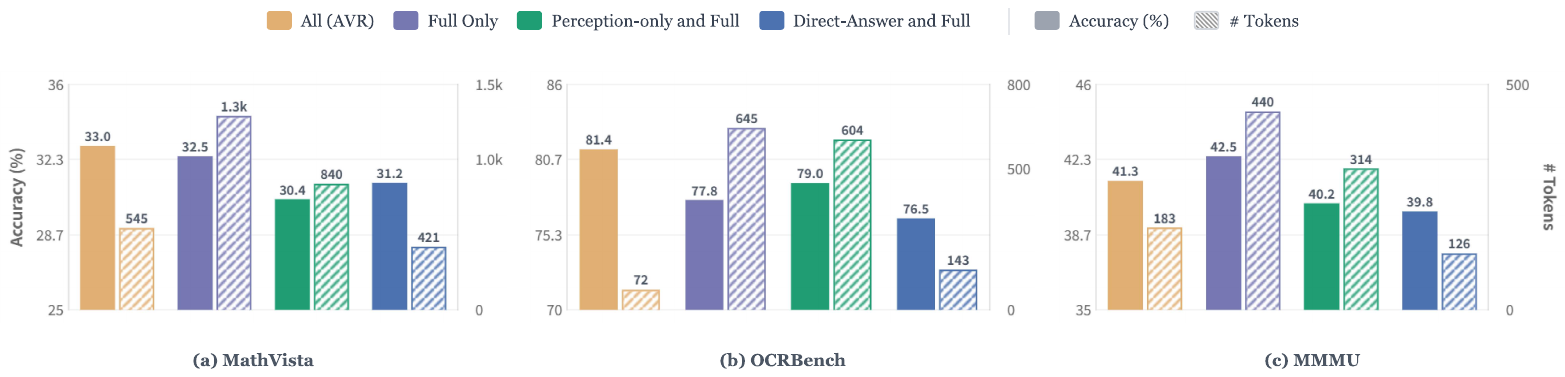}
\vspace{-5mm}
\caption{Format ablation experiment results using Qwen3-VL-2B. Except for difference in format combinations, all methods share the same hyperparameters with original method.}
\vspace{-4mm}
\label{fig:format_ablation}
\end{figure}

\subsection{Ablation on Formats}

To validate the effectiveness of our format design, we compare AVR with three alternatives: (i) full format only, (ii) perception-only + full, and (iii) direct-answer + full (Figure~\ref{fig:format_ablation}). Detailed settings are provided in Appendix~\ref{app:ablation_formats}. Using only the full format results in the highest token consumption, suggesting that excessive reasoning leads to overthinking. Adding the perception-only format improves efficiency but still requires almost 2 times the token consumption of AVR and causes a $\sim$2\% accuracy drop across benchmarks, indicating the importance of the direct-answer format for simpler queries. In contrast, combining direct answer and full format reduces tokens but lowers accuracy (1–5\%), particularly on OCRBench, implying that the lack of perception-only format harms visual grounding. Overall, the three-format design achieves the best balance between accuracy and efficiency by adaptively allocating perception, reasoning, and direct answering.

\subsection{Ablation on Diversity Reward}
\label{subsec:ablation_format_collapse}

\begin{figure}[t]
\centering
\vspace{-4mm}
\includegraphics[width=0.88\linewidth]{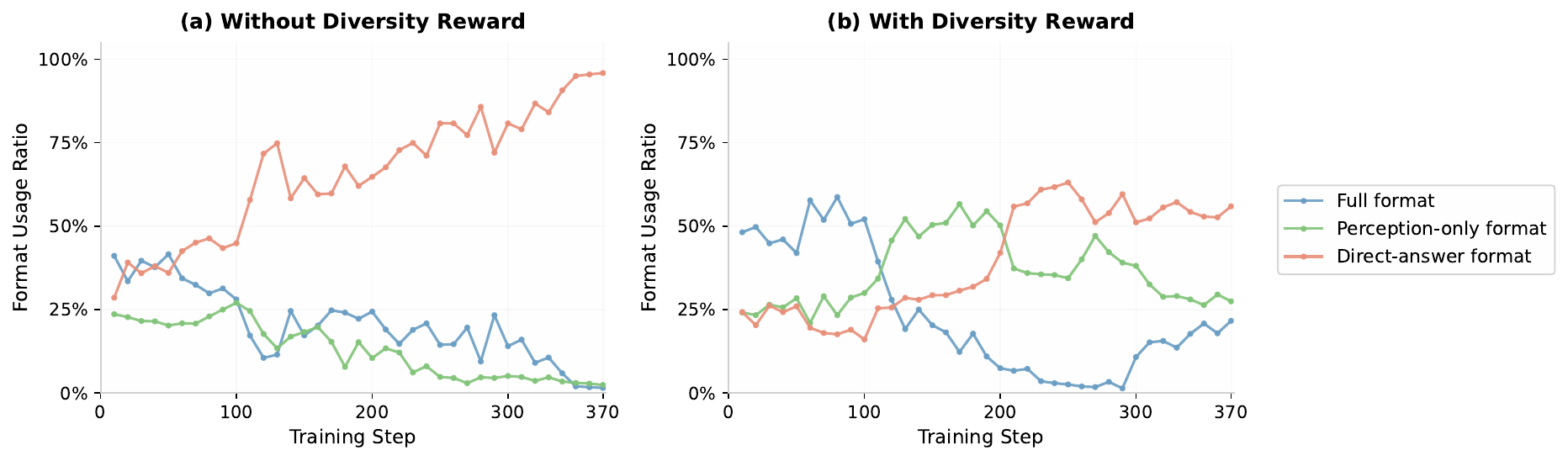}
\vspace{-4mm}
\caption{Training dynamics of response format usage during RL on Qwen3-VL-2B.
}
\label{fig:diversity_ablation}
\vspace{-2mm}
\end{figure}

To assess the effect of the diversity reward $r_{\text{div}}$, we remove it from the FS-GRPO objective while keeping all other settings unchanged (Appendix~\ref{app:ablation_diversity}). As shown in Figure~\ref{fig:diversity_ablation}, removing $r_{\text{div}}$ leads to format collapse: the model converges to a single dominant format (direct answer), while the usage of full and perception-only formats rapidly vanishes. In contrast, with $r_{\text{div}}$, the model maintains all three formats and dynamically balances their usage. These results indicate that $r_{\text{div}}$ mitigates early format dominance, enabling the model to preserve diverse reasoning strategies and learn adaptive format selection.

\subsection{Hyperparameter Sensitivity Analysis}
\label{subsec:sensitivity}

We analyze the sensitivity of AVR to two key hyperparameters: the length tolerance threshold $L$, and the format bonus $\{f_k\}$. All experiments are conducted using Qwen3-VL-2B on OCRBench and MathVista.


\begin{wraptable}{r}{0.46\linewidth}
\vspace{-12pt}
\centering
\caption{Sensitivity analysis.}
\vspace{-2mm}
\label{tab:sensitivity}

\small
\begin{minipage}{\linewidth}
\centering
\caption*{(a) Sensitivity to length tolerance $L$}
\vspace{-2mm}
\resizebox{0.75\textwidth}{!}{
\begin{tabular}{lccc}
\toprule
$L$ & OCRBench & MathVista & Tok. \\
\midrule
200  & 77.2 & 31.4 & 176.5\\
300  & 81.4 & 33.0 & 249.3\\
400  & 81.2 & 34.8 & 303.5\\
500  & 80.8 & 33.1 & 368.7\\
\bottomrule
\end{tabular}
}
\end{minipage}

\vspace{2mm}

\begin{minipage}{\linewidth}
\centering
\caption*{(b) Sensitivity to format bonus $\{f_1,f_2,f_3\}$}
\vspace{-2mm}
\resizebox{0.9\textwidth}{!}{
\begin{tabular}{lccc}
\toprule
$(f_1,f_2,f_3)$ & OCRBench & MathVista & Tok. \\
\midrule
$(0,0.1,0.2)$ & 80.7 & 32.8 & 253.2\\
$(0,0.2,0.4)$ & 81.3 & 33.2 & 275.4\\
$(0,0.3,0.4)$ & 81.0 & 33.2 & 338.5\\
$(0,0.3,0.5)$ & 81.4 & 33.0 & 249.3\\
\bottomrule
\end{tabular}
}
\end{minipage}

\vspace{-6mm}
\end{wraptable}

\vspace{-5pt}
\paragraph{Length Tolerance $L$.}
Table~\ref{tab:sensitivity}(a) reports the effect of varying $L\in \{200,300,400,500\}$. 
A small $L$ aggressively penalizes all responses, causing the model to over-compress reasoning on complex tasks and degrading accuracy. 
A large $L$ approximates no length penalty, resulting in increased token consumption without meaningful accuracy gains. 
Our default $L = 300$ achieves the best balance between efficiency and accuracy across both task types.

\vspace{-5pt}
\paragraph{Format Bonus $\{f_k\}$.}
Table~\ref{tab:sensitivity}(b) investigates the impact of varying the magnitudes and relative gaps of format bonuses. Our findings suggest that the model performance is relatively insensitive to the exact values of these parameters, provided that the ordinal relationship between the three formats is maintained. This confirms that a hierarchical bonus structure is essential for the model to reliably select the intermediate perception-only format.

Overall, these results indicate that AVR is not highly sensitive to hyperparameter choices, and its effectiveness primarily depends on preserving a proper balance between length regularization and format preferences.

\vspace{-5pt}
\section{Conclusion}
\vspace{-5pt}

We investigate overthinking in VRMs and identify Reasoning Path Redundancy as a key inefficiency driver. To address this, we propose AVR, which decomposes VRM reasoning into cognitive functions and trains models via SFT and FS-GRPO to dynamically select among three reasoning formats. Experiments across diverse benchmarks demonstrate that AVR achieves a 50--90\% token reduction while maintaining or improving accuracy, generalizing across model scales and families. Our results highlight the promise of adaptive, task-aware reasoning for more efficient multimodal inference.

\bibliography{colm2026_conference}
\bibliographystyle{colm2026_conference}

\appendix
\section{Appendix}
\subsection{Overthinking in VLM}
\label{app:overthinking}

\subsubsection{Statistics}

By analyzing 1,000 randomly sampled instances, we observed that  in Qwen3-VL-4B-Thinking, \textbf{36.7\% of the questions can be correctly answered without invoking the reasoning stage, and 32.1\% can be solved through direct answering without explicit perception or reasoning}. 
We verify this by providing the model with its generated perception output while withholding the image, or by prompting it to answer the question directly (see Box~\ref{box:perception_extra_prompt}, Box~\ref{box:perception_prompt} and Box~\ref{box:direct_prompt}). 
The perception-only and direct-answer formats are therefore designed to mitigate language reasoning overuse and compound redundancy (see Section~\ref{subsec:decoupling}).

\begin{tcolorbox}[breakable,colback=gray!5,colframe=gray!40,title=VLM Perception Extraction Prompting, label=box:perception_extra_prompt]
You are a visual perception module for a vision-language model. Your task is to analyze the image and extract only the visual information that is relevant to the question. 
Do NOT perform any reasoning or answer the question. \\[0.5em]

Your response MUST follow this format exactly: \\

\texttt{<perception>}
[Describe the visible objects, attributes, text (OCR), colors, and spatial relationships in the image that are relevant to the question.]
\texttt{</perception>} \\[0.5em]

- Focus only on what is visually observable in the image. \\
- Include objects, their attributes (color, size, text), and spatial relations if relevant. \\
- Do NOT infer hidden information. \\
- Do NOT explain reasoning. \\
- Do NOT provide the final answer. \\[0.5em]

Only output the \texttt{<perception>} block.
\end{tcolorbox}

\subsubsection{Overthinking Score Implementation}
The Overthinking Score is intended as a \textbf{diagnostic metric} to quantify the prevalence of reasoning redundancy in existing VRMs. It is not used as a training signal or evaluation criterion for our method; its sole purpose is to motivate the need for adaptive reasoning formats.
We detail the construction procedure below. Given a visual question, we first prompt the evaluated VRM Qwen3-VL-4B-Thinking to generate a response using its default reasoning mode. We then filter for correctness by comparing the model's answer against the ground-truth label via exact match. Only instances with correct original responses are retained, as redundancy is meaningful only when the original reasoning already succeeds.
For each retained response, we apply a controlled compression procedure using GPT-4o-mini. The model is instructed to rewrite the response under a minimal sufficiency constraint: removing unnecessary intermediate reasoning steps, retaining only the information essential for deriving the final answer, and producing the shortest possible response that still yields the correct answer (see the full prompt in Box~\ref{box:compression_prompt}).
After compression, we verify whether the compressed response preserves correctness by prompting GPT-4o-mini to compare the compressed answer with the ground-truth label. Instances where the compressed response is judged incorrect are discarded from the Overthinking Score computation. We compute $T_{\text{original}}$ and $T_{\text{compressed}}$ using the same tokenizer as the evaluated model to ensure consistency.

We acknowledge that exact minimality cannot be guaranteed by this procedure. However, the compressed response represents a practical upper bound on the minimal sufficient length, meaning our Overthinking Scores are likely conservative estimates of the true redundancy. To validate the reliability of this approximation, we conduct human spot-checking on a random subset of 100 compressed instances.

\begin{tcolorbox}[breakable,colback=gray!5,colframe=gray!40,
title=Overthinking Score Compression Prompt,
label=box:compression_prompt]

You are given a question and a model-generated response. Your task is to rewrite the response into a minimally sufficient answer. \\[0.5em]

Requirements: \\
- Remove redundant reasoning steps, explanations, and intermediate thoughts. \\
- Keep only the essential information required to derive the final answer. \\
- The rewritten response must remain correct. \\
- The response should be as short as possible. \\[0.5em]

Note that: \\
- Do NOT introduce new information. \\
- Do NOT change the final answer. \\
- If further shortening would make the answer incorrect or incomplete, stop. \\[0.5em]

Input:\\  
Question: \{question\}  \\
Original Response: \{response\} \\[0.5em]

Output: \\
Provide only the compressed response. 

\end{tcolorbox}

\subsubsection{Case Study}

We present several examples that are judged to exhibit overthinking.

\begin{tcolorbox}[
breakable,
colback=gray!5,
colframe=black!40,
title=Overthinking Case Study,
label=box:judge_prompt
]

\textbf{Case 1}

\vspace{0.5em}

\textbf{Question:} Of what color are the bags to the right of the man?

\vspace{0.5em}

\begin{center}
\includegraphics[width=0.25\textwidth]{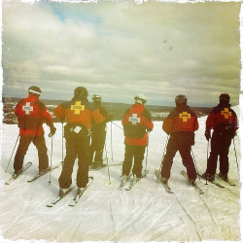}
\end{center}

\vspace{0.5em}

\textbf{Answer:}
{\ttfamily
<think> So, let's look at the image. There are several people in red jackets with yellow crosses. The question is about the bags to the right of the man. Wait, need to check each person. Let's identify the people: from left to right, the first man is on the left, then next, etc. The bags—each person has a bag on their back. Wait, the question says "the bags to the right of the man"—but which man? Wait, maybe the man in the center or a specific one. Wait, all the people are wearing red jackets with yellow crosses, and the bags on their backs—what color are the bags? Wait, no, the bags—wait, the jackets are red with yellow crosses. Wait, the bags—are they part of the jacket? Wait, no, the bags might be the backpacks. Wait, looking at the image: the people have backpacks. Let's check the color of the backpacks. The backpacks are red with yellow crosses, same as the jackets. Wait, but the question is "of what color are the bags to the right of the man?" Wait, maybe the man is one of the figures, and the bags to his right. Wait, let's count the people. There are six people. Let's take the man in the middle, but maybe the question is simpler. Wait, all the bags (backpacks) are red with yellow crosses. Wait, but the color of the bags themselves—since the jackets are red, the bags are red. Wait, let's check the image again. The people are wearing red jackets, and the backpacks are also red. So the bags (backpacks) are red. Wait, but the question says "to the right of the man"—maybe the man is one of them, so the bags on the right side. But all the bags are red. Let me confirm. Each person has a backpack that's red with a yellow cross. So the color of the bags is red.</think> \textcolor{red}{The bags (backpacks) worn by the individuals in the image are red, matching the color of their jackets, which also feature a yellow cross.}}

\vspace{0.5em}

\textbf{Analysis:} Wrong answer. Perceptual over-interpretation.

\vspace{2em}

\textbf{Case 2}

\vspace{0.5em}

\textbf{Question:} These animals are traditionally depicted as playing with a ball of what item used to make garments?

\vspace{0.5em}

\begin{center}
\includegraphics[width=0.3\textwidth]{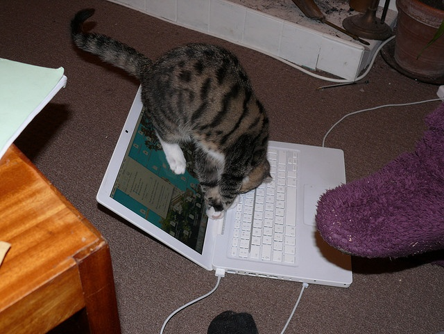}
\end{center}

\vspace{0.5em}

\textbf{Answer:}
{\ttfamily
<think>
First, observe the image. The animal visible in the picture is a cat. The cat is standing on a laptop keyboard, which indicates the animals referred to in the question are cats. Next, interpret the question. It asks: “These animals are traditionally depicted as playing with a ball of what item used to make garments?” The phrase “these animals” refers to the cats shown in the image. Now recall common cultural or cartoon depictions involving cats. In many illustrations, cartoons, and children's stories, cats are frequently shown playing with a ball of yarn. The yarn is usually rolled into a ball, which cats bat around and unravel while playing. Then consider the second part of the question: the item is something “used to make garments.” Yarn fits this description because it is commonly used in knitting or crocheting to create clothing items such as sweaters, scarves, hats, and socks. Putting these observations together: cats are traditionally depicted playing with balls of yarn, and yarn is a textile material used to produce garments. Therefore, the item referred to in the question is yarn.
</think> \textcolor{red}{The answer is yarn.}}

\vspace{0.5em}

\textbf{Analysis:} Correct answer. Language reasoning overuse.

\vspace{2em}

\textbf{Case 3}

\vspace{0.5em}

\textbf{Question:} $ABCD$ is a rhombus. If $PB = 12$, $AB = 15$, and $\angle ABD = 24$, Find $\angle BDA$.

\vspace{0.5em}

\begin{center}
\includegraphics[width=0.3\textwidth]{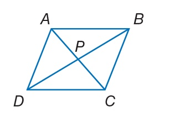}
\end{center}

\vspace{0.5em}

\textbf{Answer:}
{\ttfamily
<think>
Got it, let's try to solve this problem. First, let's recall what a rhombus is. A rhombus is a quadrilateral with all sides equal, and its diagonals bisect each other at right angles? Wait, no, not necessarily right angles, but they bisect the angles and bisect each other. Wait, actually, in a rhombus, the diagonals bisect the vertex angles, and they are perpendicular bisectors of each other. Wait, is that true? Let me confirm: yes, in a rhombus, the diagonals are perpendicular bisectors of each other. So diagonals AC and BD intersect at P, so AP = PC, BP = PD, and angle APB is 90 degrees. Wait, but the problem says PB = 12, so since diagonals bisect each other, PD = PB = 12? Wait, no, in a rhombus, diagonals bisect each other, so yes, BP = PD. Wait, but maybe I need to check. Wait, the problem says: ABCD is a rhombus, so AB = BC = CD = DA = 15 (since AB = 15). PB = 12, so BD = BP + PD = 24 (since diagonals bisect each other, so PD = PB = 12). Now, angle ABD = 24 degrees. Let's look at triangle ABD: in a rhombus, AB = AD (all sides equal), so triangle ABD is isosceles with AB = AD. Wait, AB = AD, so angles opposite them are equal. Angle ABD is the angle at B, which is angle ABD = 24 degrees, so angle BDA is the angle at D, which should be equal to angle ABD if AB = AD. Wait, is that right? Wait, let's clarify: in triangle ABD, vertices A, B, D. Sides: AB = AD (since ABCD is a rhombus), so yes, triangle ABD is isosceles with AB = AD. Therefore, angles opposite the equal sides are equal. Side opposite angle ABD (which is at B) is AD, and side opposite angle BDA (which is at D) is AB. Since AB = AD, angles at B and D are equal. So angle ABD = angle BDA = 24 degrees. Wait, is that correct? Let's verify with the diagonals. In a rhombus, the diagonals bisect the angles. So angle ABD is half of angle ABC, but wait, angle ABD is given as 24 degrees, so angle ABC = 48 degrees. But in a rhombus, opposite angles are equal, and adjacent angles are supplementary. But maybe we don't need that. Wait, another way: in triangle ABD, AB = AD = 15, so it's isosceles with base BD. Therefore, base angles at B and D are equal. Angle at B is angle ABD = 24 degrees, so angle at D is angle ADB = 24 degrees. That seems right. Wait, but let's make sure we're not confusing the angles. Let's label the rhombus: vertices in order A, B, C, D, so sides AB, BC, CD, DA. So diagonal BD connects B to D, and diagonal AC connects A to C, intersecting at P. $\angle ABD$ is the angle at B between AB and BD. So in triangle ABD, angle at B is angle ABD = 24°, angle at D is angle ADB (which is what we need, $\angle BDA$), and angle at A is angle BAD. Since AB = AD (rhombus sides), triangle ABD is isosceles with AB = AD, so angles at B and D are equal. Therefore, angle ABD = angle ADB = 24°. Wait, is there any chance that AB is not equal to AD? No, because in a rhombus, all sides are equal, so AB = BC = CD = DA. Let's check with the diagonals. In a rhombus, diagonals bisect each other, so BP = PD = 12 (since PB = 12). Now, in triangle ABP, we have AB = 15, BP = 12, and angle ABP = 24°. But we don't need to calculate anything else, because we can use the isosceles triangle property. Wait, let's confirm with the Law of Sines in triangle ABD. Law of Sines says AB / sin(angle ADB) = AD / sin(angle ABD). Since AB = AD, sin(angle ADB) = sin(angle ABD), so angle ADB = angle ABD.
</think> \textcolor{red}{The answer should be: $\angle BDA$ = 24°.}}

\vspace{0.5em}

\textbf{Analysis:} Correct answer. Compound redundancy.

\end{tcolorbox}

\subsection{Dataset Composition}
\label{app:dataset_composition}

Table~\ref{tab:training_data} presents the composition of our RL training dataset.
The dataset contains 44k samples aggregated from 10 vision-language datasets covering four task categories: perception-centric tasks, compositional visual reasoning, scientific and mathematical reasoning, and general knowledge VQA.
These categories span different reasoning requirements, providing a diverse training distribution for learning adaptive reasoning formats.

\begin{table}[h]
\centering
\small
\begin{tabular}{lccc}
\toprule
\textbf{Category} & \textbf{Representative Datasets} & \textbf{Samples} & \textbf{Ratio} \\
\midrule
Perception-Centric Tasks 
& OCR-VQA, ChartQA, GQA 
& 11k 
& 25.0\% \\

Compositional Visual Reasoning 
& CLEVR, VCR 
& 9k 
& 20.5\% \\

Scientific \& Mathematical Reasoning 
& MathVerse, Geometry3K, ScienceQA 
& 9k 
& 20.5\% \\

General Knowledge VQA 
& OK-VQA, TQA 
& 15k 
& 34.0\% \\

\midrule
\textbf{Total} & 10 datasets & 44k & 100\% \\
\bottomrule
\end{tabular}
\caption{Composition of the RL training dataset. 
This diverse distribution enables the model to learn adaptive reasoning formats across tasks with different reasoning requirements.}
\label{tab:training_data}
\end{table}

\subsection{Format Construction}
\label{app:format_construction}
The three reasoning formats are designed to correspond to the minimal functional decomposition of visual reasoning in vision-language models. In most VQA-style tasks, solving a problem typically involves three conceptual stages: perception, reasoning, and answer generation. However, not all questions require the full pipeline. Some tasks can be solved directly from prior knowledge or simple visual recognition, similar to how humans can answer certain questions without explicit reasoning.

Based on this observation, we define three response formats with increasing computational depth: Direct Answer, Perception-Only, and Full Format. These formats correspond respectively to using none, a subset, or the complete reasoning pipeline. This design forms a minimal basis for representing different reasoning pathways while keeping the action space compact enough for stable reinforcement learning. In contrast, using fewer formats (e.g., full reasoning only) fails to explicitly capture visual information, which may lead to reasoning based on incomplete evidence. Introducing more formats would unnecessarily enlarge the decision space without providing additional functional benefit. Therefore, the three-format design provides a principled and efficient decomposition of visual reasoning behaviors.


\subsection{SFT Dataset Construction}
\label{app:sft_data}
To construct the multi-format SFT dataset, we processed 44k image-text pairs through a three-step pipeline: response generation, functional format verification, and quality filtering.

\paragraph{Step 1: Response Generation.}
We prompt GPT-4o~\citep{hurst2024gpt} to generate responses following our decoupled reasoning architecture consisting of \texttt{<perception>}, \texttt{<reasoning>}, and \texttt{<answer>} (see Box~\ref{box:reasoning_prompt}). Responses whose final answers do not match the ground-truth label (via exact match) are discarded.

\paragraph{Step 2: Functional Format Verification.}
We emphasize that format assignment is not based on GPT-4o's subjective judgment about whether reasoning is ``needed.'' Instead, we apply functional tests that rely solely on answer correctness:

\begin{itemize}
    \item \textbf{Perception-Only Test.} We provide a separate model with only the generated \texttt{<perception>} text and the original question, \emph{withholding the image entirely}. If the model consistently derives the correct answer from this textual description alone, the reasoning stage is functionally redundant for this instance, and the sample is assigned the perception-only format (see Box~\ref{box:perception_prompt}).
    \item \textbf{Direct Answer Test.} We prompt the model with the image and question but instruct it to answer directly without intermediate reasoning. If the correct answer is produced, the sample is assigned the direct answer format (see Box~\ref{box:direct_prompt}).
    \item \textbf{Full Format.} Samples that fail both tests above---i.e., require explicit reasoning beyond perception to reach the correct answer---are retained in the full decoupled format containing \texttt{<perception>}, \texttt{<reasoning>}, and \texttt{<answer>}.
\end{itemize}

This verification logic is self-contained: it depends only on whether the correct answer can be recovered under a restricted input condition, not on any model's opinion about task difficulty.

\paragraph{Step 3: Quality Filtering.}
After format assignment, we perform additional filtering including format distillation and human inspection. The final SFT dataset contains 11k high-quality samples distributed across the three formats.

\paragraph{Note on SFT Label Sensitivity.}
We note that the SFT stage serves as a warm start to teach the model format competency, the ability to generate well-formed responses in all three formats, instead of optimal format policy. The selection of which format to use for a given task is learned during the subsequent RL stage, where the reward signal is based on verifiable answer correctness, not on SFT labels. Therefore, even if some SFT format labels are noisy, the RL stage can correct suboptimal format assignments through its own optimization. This design makes the overall framework robust to moderate label noise in the SFT data.

\begin{tcolorbox}[breakable,colback=gray!5,colframe=gray!40,title=Reasoning Decomposition Prompting, label=box:reasoning_prompt]

You are an expert visual reasoning assistant. Your task is to analyze images and solve problems by decomposing your thought process into a strictly decoupled structure.\\[0.5em]

Your response MUST follow this exact format without any preamble or conversational filler: \\
\texttt{<perception>}
[Detailed visual grounding and observation of relevant entities/attributes in the image]
\texttt{</perception>} 
\texttt{<reasoning>}
[Step-by-step logical deduction based ON the perception part to reach the final conclusion]
\texttt{</reasoning>} 
\texttt{<answer>}
[The final concise answer to the question]
\texttt{</answer>} \\[0.5em]

1. \texttt{<perception>}: Focus only on WHAT you see. Describe key objects, their spatial relationships, text/OCR, and colors relevant to the user's question. Act as a "vision-to-text" module. \\
2. \texttt{<reasoning>}: Do not repeat observations. Instead, use the facts from the \texttt{<perception>} section to solve the logic puzzle or answer the question. \\
3. \texttt{<answer>}: Provide only the final result (e.g., a single word, a number, or a short phrase). \\[0.5em]

- Do not use phrases like "Based on the image" or "In this picture" inside the tags. \\
- Ensure the \texttt{<perception>} part contains enough information so that a human could theoretically answer the question WITHOUT seeing the image. \\
- Strictly adhere to the XML-style tags.
\end{tcolorbox}

\begin{tcolorbox}[breakable,colback=gray!5,colframe=gray!40,
title=Perception-Only Verification Prompt,
label=box:perception_prompt]

You are given a visual description extracted from an image, together with a question about the image. 
Your task is to answer the question using ONLY the provided description. 
You do NOT have access to the original image. \\[0.5em]

Visual description:
\texttt{<perception>}
[Generated perception text]
\texttt{</perception>} \\
Question:
[User question] \\[0.5em]

- Answer the question based solely on the information contained in the perception text. \\
- Do NOT assume any visual details that are not explicitly described. \\
- If the perception text contains sufficient information, output the correct answer. \\
- Provide only the final answer without additional reasoning or explanation. \\[0.5em]

Output format:
\texttt{<answer>}
[Final answer]
\texttt{</answer>}

\end{tcolorbox}

\begin{tcolorbox}[breakable,colback=gray!5,colframe=gray!40,
title=Direct Answer Prompt,
label=box:direct_prompt]

You are a visual question answering assistant. 
Given an image and a question, your task is to provide the answer as directly and concisely as possible. \\[0.5em]

- Carefully observe the image and read the question. \\
- If the answer is immediately apparent, respond directly with the final answer. \\
- Do NOT produce intermediate reasoning steps. \\
- Do NOT describe the image unless necessary. \\[0.5em]

Output format:
\texttt{<answer>}
[Final concise answer]
\texttt{</answer>}

\end{tcolorbox}

\subsection{Implementation Details}
\label{app:implementation_details}

\subsubsection{Stage 1: SFT}
SFT is implemented using the LLaMA-Factory~\citep{zheng2024llamafactory} framework. We freeze the vision encoder and the multimodal projector, and only update the language model parameters~\citep{liu2023visual} with full fine-tuning. This design preserves the pretrained visual representations while allowing the language model to adapt to the response format supervision.

Training is performed with a per-device batch size of 4 and gradient accumulation of 2 (i.e., effective batch size of 8), using the AdamW optimizer~\citep{loshchilov2017decoupled} with a learning rate of $5\times10^{-5}$ and cosine learning rate scheduling with a warmup ratio of 0.1~\citep{loshchilov2016sgdr}. We train the model for one epoch with DeepSpeed ZeRO-3 optimization~\citep{rajbhandari2020zero} to enhance efficiency. The maximum image resolution is limited to $262{,}144$ pixels to control memory usage.

\subsubsection{Stage 2: RL}
\label{app:grpo_setting}

RL training is implemented using the VeRL~\citep{sheng2025hybridflow} framework under the FS-GRPO objective described in Section~\ref{subsec:rl_reward}. The SFT model initializes the actor policy, while a frozen reference model is used for KL regularization.

\paragraph{Training Configurations}
For each prompt, the model samples $G=8$ responses to form a GRPO group. Responses are generated using nucleus sampling with temperature $0.8$ and top-$p=0.95$~\citep{holtzman2019curious} with a maximum length of 2048 tokens. Training uses a batch size of 128 prompts and a validation batch size of 32, with 10\% of the data reserved for validation. Policy optimization uses a learning rate of $5\times10^{-7}$ with PPO clipping ratio $0.2$~\citep{schulman2017proximal}. KL regularization with coefficient $0.02$~\citep{ouyang2022training} is applied to stabilize policy updates.

\paragraph{Reward Design}
The format-aware reward function used in RL training is described in Section~\ref{subsec:rl_reward}. In our implementation, the format bonuses encourage concise reasoning by assigning larger rewards to more efficient formats (direct answer $>$ perception-only $>$ full reasoning). Specifically, we set $f_1=0$, $f_2=0.3$, and $f_3=0.5$. To discourage excessively long responses, the length scaling factor uses a token tolerance threshold of $L=300$, the average length of SFT data responses. In addition, the diversity reward follows the cosine decay schedule described in Section~\ref{subsec:rl_reward}.

All training is conducted using eight NVIDIA A6000 GPUs.

\subsection{FS-GRPO Training Dynamics}
\label{app:grpo_details}

\begin{figure}[t]
\centering
\includegraphics[width=0.9\linewidth]{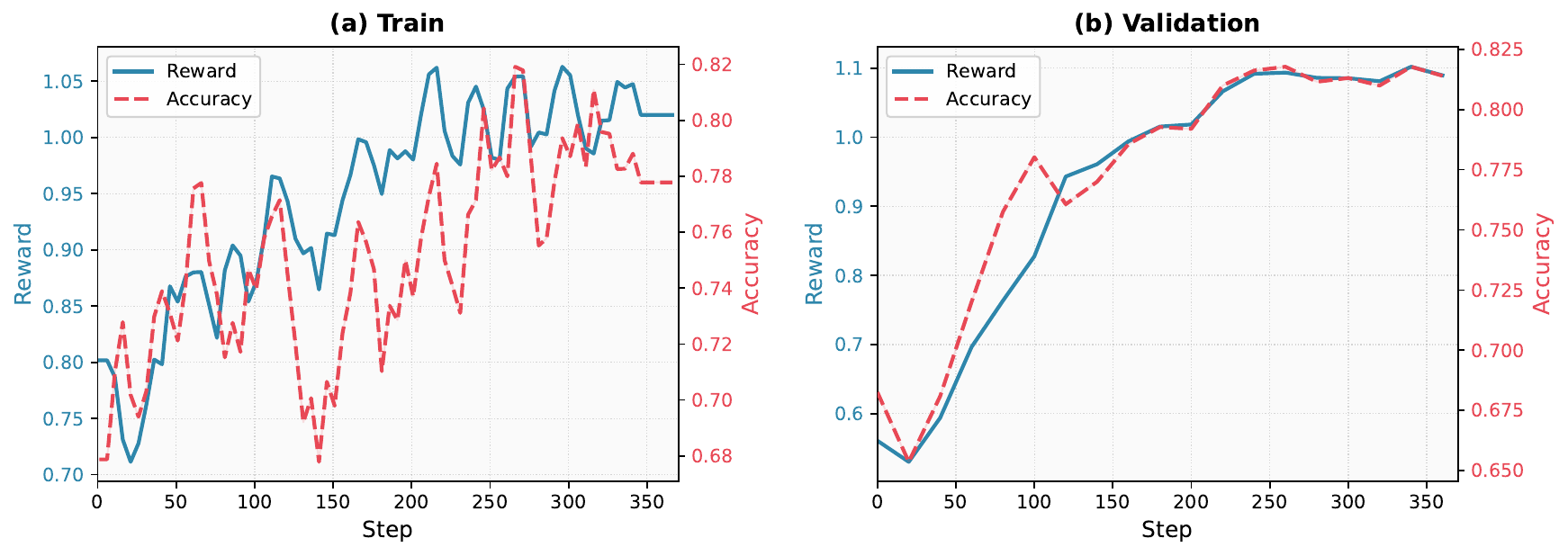}
\caption{
Training dynamics of FS-GRPO optimization on Qwen3-VL-2B. We report the mean reward and answer accuracy on both the training and validation sets.
}
\label{fig:training_dynamics}
\end{figure}

To further analyze the behavior of FS-GRPO optimization, we track the mean reward and answer accuracy during RL training.
Figure~\ref{fig:training_dynamics} shows the training and validation curves.

Both reward and accuracy steadily increase throughout training, indicating that the policy gradually improves while maintaining stable optimization. The consistent improvement of both reward and accuracy suggests that the reward design is well aligned with the target objective. 

\begin{wrapfigure}{r}{0.47\textwidth}
\vspace{-15pt}
\centering
\includegraphics[width=\linewidth]{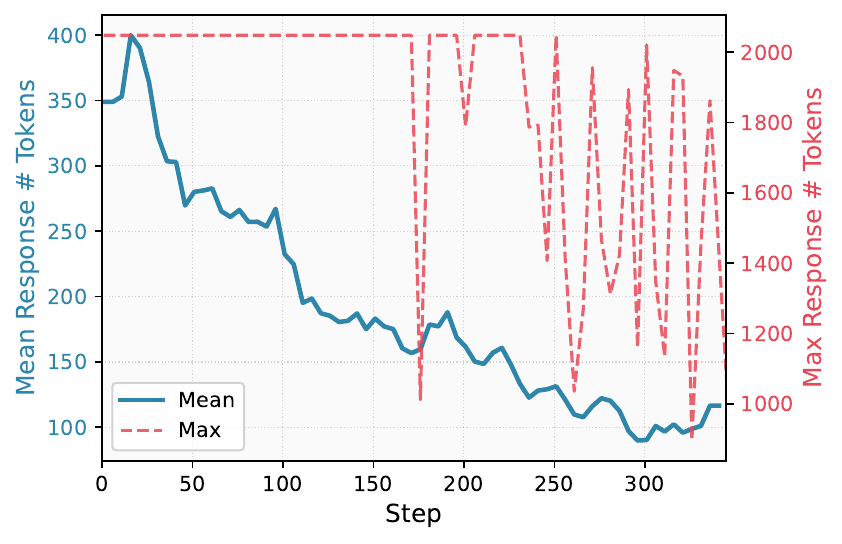}
\caption{
Response length during FS-GRPO on Qwen3-VL-2B. 
We report the mean and maximum number of generated tokens per response across training steps.
}
\label{fig:training_length}
\vspace{-40pt}
\end{wrapfigure}

Additionally, Figure~\ref{fig:training_length} shows the response length dynamics during training. The mean number of generated tokens gradually decreases over training steps, indicating improved generation efficiency as the policy becomes more confident in producing concise responses, while the maximum length exhibits larger fluctuations. 

In addition, the format usage dynamics during training are shown in Figure~\ref{fig:diversity_ablation} in the main text. That figure illustrates how the diversity reward prevents format collapse and enables the model to maintain a balanced mixture of response formats during learning.

On the validation set, both reward and accuracy follow a similar upward trend and stabilize in later stages of training. This indicates that the learned policy generalizes well and that the reward design effectively guides the model toward more efficient reasoning strategies without overfitting.

\subsection{Evaluation Details}

During inference, we set the sampling temperature to 0.8, top-$p$ to 1.0, and the maximum response length to 8096 tokens to avoid truncation for long reasoning outputs. 

For answer evaluation, we adopt different matching strategies depending on the question format. For multiple-choice questions, we apply structured answer extraction to parse the final predicted option and compare it with the ground-truth label. For open-ended questions, where answers may vary in expression, we employ Gemini-2.5-Flash as an automatic judge to determine correctness. The judge model evaluates semantic equivalence between the predicted answer and the reference answer following a standardized evaluation prompt set in VLMEvalKit.

This evaluation protocol ensures reliable and consistent comparison across different benchmarks and answer formats.

\subsection{Generalization Across Model Families}
\label{app:other_model_family}

To evaluate the generality of our approach beyond the Qwen3-VL family, we conduct additional experiments on the InternVL3 series~\citep{zhu2025internvl3}. InternVL3 adopts a different multimodal paradigm (ViT-MLP-LLM) with a stronger vision encoder and a projector-based fusion mechanism, providing a complementary testbed for assessing the robustness of our method across heterogeneous model families.

We consider two variants from the InternVL3 family:
(1) \textbf{InternVL3-Instruct}, which is trained with SFT only, and
(2) \textbf{InternVL3}, which further incorporates preference optimization (RL) on top of SFT.

While InternVL3 does not provide explicitly "thinking" variants, the combination of SFT and preference optimization encourages the model to generate structured reasoning traces. In practice, prompting the model with "think step by step" consistently induces long-form CoT reasoning~\citep{kojima2022large}.

On top of InternVL3-Instruct, we apply the same training pipeline as described in Section~\ref{sec:method}. We compare the following variants:
\begin{itemize}
    \item \textbf{CoT}: always-thinking baseline, representing standard CoT usage;
    \item \textbf{Instruct}: original instruction-tuned model;
    \item \textbf{+ SFT}: multi-format supervised fine-tuning;
    \item \textbf{AVR}: our full method with format-aware RL optimization.
\end{itemize}

\subsubsection{Results}
Table~\ref{tab:internvl3_results} presents the performance across reasoning-intensive, perception-intensive, and general benchmarks.

We observe consistent trends with the Qwen3-VL family. AVR significantly reduces token usage across all task categories, especially on OCRBench, representative perception-intensive benchmark. Despite shorter responses, AVR maintains or improves accuracy in most settings. The model selectively allocates reasoning effort, using shorter formats for simple tasks while preserving full reasoning for complex ones.

Compared to the CoT baseline, AVR achieves substantial reductions in token consumption while avoiding the overthinking behavior induced by unconditional CoT generation.

\subsubsection{Analysis}
These results demonstrate that our approach is not tied to a specific architecture or training paradigm. InternVL3 differs from Qwen3-VL in both multimodal fusion (projector-based vs. native fusion) and training strategy (separate vision encoder and LLM with post-hoc alignment), yet AVR achieves consistent improvements.

Overall, the cross-family results suggest that adaptive reasoning is a general capability that can be learned on top of diverse multimodal models with emergent CoT abilities, rather than relying on specific architectural or training choices.

\begin{table}[t]
\centering
\small
\resizebox{0.6\textwidth}{!}{
\begin{tabular}{llcccccc}
\toprule
\multirow{2}{*}{Size} & \multirow{2}{*}{Variant}
& \multicolumn{2}{c}{MathVista}
& \multicolumn{2}{c}{OCRBench}
& \multicolumn{2}{c}{MMMU} \\

\cmidrule(lr){3-4}
\cmidrule(lr){5-6}
\cmidrule(lr){7-8}

& 
& Acc & Tok.
& Acc & Tok.
& Acc & Tok. \\

\midrule

\multirow{5}{*}{1B}
& CoT & \second{41.8} & 1329.3 & 73.4 & 453.7 & \best{40.4} & 829.3 \\
& Instruct        & 32.4 & \second{674.1} & 75.3 & 298.4 & 33.5 & 604.5 \\
& + SFT           & 37.2 & 798.3 & \second{76.8} & \second{192.3} & 37.2 & \second{532.8} \\
\rowcolor{gray!12}
& \textbf{AVR}    & \best{42.3} & \best{645.6} & \best{77.6} & \best{63.9} & \second{40.0} & \best{374.9} \\
& $\Delta$        & +0.5 & -51.4\% & +4.2 & -85.9\% & -0.4 & -54.8\% \\

\midrule

\multirow{5}{*}{2B}
& CoT & \best{48.1} & 1657.4 & 77.2 & 531.8 & \best{44.8} & 980.6 \\
& Instruct        & 41.3 & 869.2 & \second{77.9} & 210.6 & 38.1 & 748.5 \\
& + SFT           & 40.8 & \second{782.3} & 76.3 & \second{156.3} & 37.4 & \second{687.3} \\
\rowcolor{gray!12}
& \textbf{AVR}    & \second{47.2} & \best{735.2} & \best{79.6} & \best{84.7} & \second{43.2} & \best{421.0} \\
& $\Delta$        & -0.9 & -55.6\% & +2.4 & -84.1\% & -1.6 & -57.1\% \\

\bottomrule
\end{tabular}
}
\caption{
Cross-family evaluation on InternVL3 models.
$\Delta$ is computed with respect to the CoT baseline. Accuracy denotes absolute difference, and token reduction denotes relative percentage decrease.
}
\label{tab:internvl3_results}
\end{table}

\subsection{Details on Format Selection Across Tasks}
\label{app:fotmat_selection}

\begin{table}[t]
\centering
\small
\resizebox{0.8\textwidth}{!}{
\begin{tabular}{lcccccc}
\toprule
\multirow{2}{*}{\normalsize Benchmark} & 
\multicolumn{2}{c}{\normalsize Direct Answer} & 
\multicolumn{2}{c}{\normalsize Perception-only Format} & 
\multicolumn{2}{c}{\normalsize Full Format} \\
\cmidrule(lr){2-3} \cmidrule(lr){4-5} \cmidrule(lr){6-7}
& Ratio & Acc & Ratio & Acc & Ratio & Acc \\
\midrule
OCRBench & 29.4 & 86.2 & 47.5 & 90.3 & 23.1 & 75.2 \\
MathVista & 5.9 & 35.4 & 15.8 & 42.6 & 78.3 & 43.7 \\
MMMU & 12.0 & 55.6 & 33.7 & 59.2 & 54.3 & 61.1 \\
\bottomrule
\end{tabular}
}
\caption{
Format selection ratios (\%) and corresponding accuracies across different benchmarks.}
\vspace{-10pt}
\label{tab:detailes_format_selection}
\end{table}

Table~\ref{tab:detailes_format_selection} presents a detailed breakdown of format usage and corresponding accuracy across three representative benchmarks. We observe that the preferred reasoning format strongly correlates with the intrinsic task requirements.

\vspace{-5pt}
\paragraph{Perception-Intensive Tasks.}
On OCRBench, the model predominantly adopts the perception-only format (47.5\%), which also achieves the highest accuracy (90.3\%). In contrast, full format achieves only 75.2\%, suggesting that introducing unnecessary reasoning may harm performance due to error accumulation. Direct answers also remain competitive (86.2\%), indicating that many instances can be solved with minimal reasoning.

\vspace{-5pt}
\paragraph{Reasoning-Intensive Tasks.}
On MathVista, the full format dominates (78.3\%) and yields the best accuracy (43.7\%), confirming that complex tasks benefit from explicit reasoning. Lightweight formats such as direct answer are rarely selected and perform significantly worse, demonstrating that insufficient reasoning leads to suboptimal results.

\vspace{-5pt}
\paragraph{General Tasks.}
On MMMU, the model adopts a more balanced strategy, distributing predictions across all three formats. Except from full format, but both perception-only (59.2\%) and direct answer (55.6\%) remain competitive. This suggests that general-domain tasks exhibit heterogeneous reasoning requirements, where adaptive format selection becomes particularly beneficial.

Overall, these results highlight that no single reasoning format is universally optimal. Instead, effective performance arises from dynamically matching the reasoning format to task complexity: lightweight formats are preferred for perception-heavy problems, while full reasoning is reserved for tasks requiring multi-step inference. This validates the design of AVR in enabling adaptive reasoning path selection.

\subsection{Ablation Details}

\subsubsection{Ablation on Formats}
\label{app:ablation_formats}
In the three alternatives—(i) full format only, (ii) perception-only together with full format, and (iii) direct-answer together with full format—the reward formulation follows Equation~\ref{eq:reward_decomposition}. For the full-format-only setting, no format bonus is applied. For the other two alternatives, the format bonuses remain identical to those used in the main setting, as specified in Appendix~\ref{app:grpo_setting}.

\subsubsection{Ablation on Diversity Reward}
\label{app:ablation_diversity}

In the ablation setting where the diversity reward is removed, the total reward $r$ for each response is defined as
\begin{equation}
    r = s_{L} \cdot r_{\text{format}}.
\end{equation}
Here, $r_{\text{format}}$ encourages correct answers with efficient formats, while $s_{L}$ penalizes overly long responses. 
All other reward hyperparameters are kept identical to the original formulation described in Appendix~\ref{app:grpo_setting}.

\subsection{Case Study}
\label{app:case_study}

\subsubsection{Qualitative Examples of Adaptive Reasoning}
We present representative examples generated by AVR-4B to analyze how the model adapts its reasoning format across tasks with varying complexity.

\begin{tcolorbox}[
breakable,
colback=gray!5,
colframe=black!40,
title=AVR Case Study,
label=box:avr_case_study
]

\textbf{Case 1}

\vspace{0.5em}

\textbf{Question:} Of what color are the bags to the right of the man?

\vspace{0.5em}

\begin{center}
\includegraphics[width=0.25\textwidth]{images/what_color.png}
\end{center}

\vspace{0.5em}

\textbf{Answer:}
{\ttfamily
\textcolor{red}{<answer>Black</answer>}}

\vspace{0.5em}

\textbf{Analysis:} Correct answer. Direct answer.

\vspace{2em}

\textbf{Case 2}

\vspace{0.5em}

\textbf{Question:} What does the sign say right above the stop sign?

\vspace{0.5em}

\begin{center}
\includegraphics[width=0.3\textwidth]{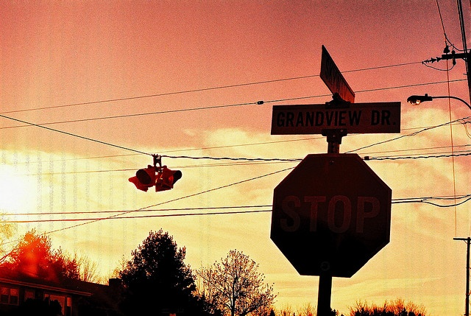}
\end{center}

\vspace{0.5em}

\textbf{Answer:}
{\ttfamily
\textcolor{green}{<perception>The image shows a stop sign with another street sign mounted directly above it. The text on that upper sign reads "GRANDVIEW DR."</perception>}\textcolor{red}{<answer>GRANDVIEW DR</answer>}}

\vspace{0.5em}

\textbf{Analysis:} Correct answer. Percception-only format.

\vspace{2em}

\vspace{2em}

\textbf{Case 3}

\vspace{0.5em}

\textbf{Question:} When is the sign active?

\vspace{0.5em}

\begin{center}
\includegraphics[width=0.3\textwidth]{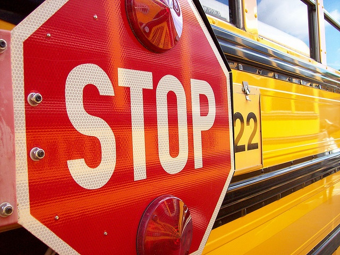}
\end{center}

\vspace{0.5em}

\textbf{Answer:}
{\ttfamily
\textcolor{green}{<perception>The image shows a school bus with a red STOP sign extended on its side,} \textcolor{blue}{which is used to control traffic when children are getting on or off the bus.}\textcolor{green}{</perception>}\textcolor{red}{<answer>When children are boarding or leaving the school bus</answer>}}

\vspace{0.5em}

\textbf{Analysis:} Correct answer. Perception-only format. The perception block includes a brief functional interpretation (“which is used to control traffic...”), which goes beyond strictly observable visual content.

\vspace{2em}

\textbf{Case 4}

\vspace{0.5em}

\textbf{Question:} These animals are traditionally depicted as playing with a ball of what item used to make garments?

\vspace{0.5em}

\begin{center}
\includegraphics[width=0.3\textwidth]{images/these_animals.png}
\end{center}

\vspace{0.5em}

\textbf{Answer:}
{\ttfamily
\textcolor{green}{<perception>The image shows a cat standing on a laptop keyboard.</perception>}\textcolor{blue}{<reasoning>Cats are commonly depicted playing with balls of yarn, which is used to make garments. </reasoning>}\textcolor{red}{<answer>yarn</answer>}}

\vspace{0.5em}

\textbf{Analysis:} Correct answer. Full format.

\vspace{2em}

\textbf{Case 5}

\vspace{0.5em}

\textbf{Question:} $ABCD$ is a rhombus. If $PB = 12$, $AB = 15$, and $\angle ABD = 24$, Find $\angle BDA$.

\vspace{0.5em}

\begin{center}
\includegraphics[width=0.3\textwidth]{images/ABCD.png}
\end{center}

\vspace{0.5em}

\textbf{Answer:}
{\ttfamily
\textcolor{green}{<perception>The figure shows a rhombus $ABCD$ with diagonals intersecting at $P$. Given $PB = 12$, so diagonal $BD = 24$. All sides of the rhombus are equal, so $AB = AD = 15$.</perception>}\textcolor{blue}{<reasoning> Consider triangle $ABD$. Since $AB = AD$, it is an isosceles triangle with base $BD$. Therefore, the base angles are equal: $\angle ABD = \angle BDA$. Given $\angle ABD = 24^\circ$, we have $\angle BDA = 24^\circ$.</reasoning>}\textcolor{red}{<answer>$24^\circ$</answer>}}

\vspace{0.5em}

\textbf{Analysis:} Correct answer. Full format.

\end{tcolorbox}

\subsubsection{Reasoning Path Robustness}

A potential concern is that the model may superficially conform to the predefined formats while implicitly performing hidden reasoning within earlier stages (e.g., embedding reasoning inside the perception block, as shown in Case 3 of Box~\ref{box:avr_case_study}). Such behavior would undermine the intended decomposition and reduce the interpretability of the framework.

To evaluate this, we conduct a manual inspection of 1000 randomly sampled evaluation instances and examine whether stage boundaries are violated. We find that only 4.3\% of cases exhibit any form of stage mixing, indicating that the model largely adheres to the intended functional separation between perception, reasoning, and answer generation.

This result suggests that the learned format control is not merely superficial, but reflects a genuine structural decomposition of the reasoning process. The low rate of stage violation provides evidence that AVR enforces meaningful modularization rather than format-level mimicry.

\end{document}